\documentclass[letterpaper, 10 pt, conference]{ieeeconf}  

\IEEEoverridecommandlockouts                              

\usepackage{multirow}
\usepackage{amsmath}  
\usepackage{amssymb}  
\usepackage[colorlinks,urlcolor=blue,linkcolor=blue,citecolor=blue]{hyperref}
\usepackage{color,array}
\usepackage{graphicx}
\usepackage{cite}
\usepackage{setspace}  
\usepackage{subfigure}  
\usepackage{graphics} 
\usepackage{epsfig} 
\usepackage{times} 
\usepackage{amsmath} 
\usepackage{amssymb}  
\usepackage{subfigure}

\usepackage[linesnumbered,ruled,vlined]{algorithm2e}

\usepackage{multirow} 
\usepackage{booktabs} 
\usepackage{gensymb}  
\usepackage{bm}  
\usepackage{underscore}  

\usepackage{amsmath} 
\usepackage{amssymb}  
\usepackage[colorlinks,linkcolor=blue]{hyperref}  
\title{\LARGE \bf
Following Closely: A Robust Monocular Person Following System for Mobile Robot
}

\author{Hanjing Ye$^{1}$, Jieting Zhao$^{2}$, Yaling Pan$^{1}$, Weinan Chen$^{2}$ and Hong Zhang$^{2*}$
\thanks{*corresponding author (hzhang@sustech.edu.cn).}
\thanks{$^{1}$Hanjing Ye and Yaling Pan are with the Biomimetic and Intelligent Robotics Lab (BIRL), Guangdong
University of Technology, Guangzhou, China.}%
\thanks{$^{2}$Jieting Zhao, Weinan Chen and Hong Zhang are with the Department of Electronic and Electrical Engineering, Southern University of Science and Technology, Shenzhen, China.}%
}

\begin{document}

\maketitle
\thispagestyle{empty}
\pagestyle{empty}

\begin{abstract}
Monocular person following (MPF) is a capability that supports many useful applications of a mobile robot. However, existing MPF solutions are not completely satisfactory. Firstly, they often fail to track the target at a close distance either because they are based on visual servo or they need the observation of the full body by the robot. Secondly, their target Re-IDentification (Re-ID) abilities are weak in cases of target appearance change and highly similar appearance of distracting people. To remove the assumption of full-body observation, we propose a \textit{width-based} tracking module, which relies on the target width, which can be observed even at a close distance. For handling issues related to appearance variation, we use a global CNN (convolutional neural network) descriptor to represent the target and a ridge regression model to learn a target appearance model online. We adopt a sampling strategy for online classifier learning, in which both long-term and short-term samples are involved. We evaluate our method in two datasets including a public person following dataset and a custom-built with challenging target appearance and target distance. Our method achieves state-of-the-art (SOTA) results on both datasets. The code and dataset of our work in this research are publicly available in \href{https://github.com/MedlarTea/MPF_GRR_SLT}{https://github.com/MedlarTea/MPF_GRR_SLT}.
\end{abstract}

\section{INTRODUCTION}
Nowadays, mobile robotics is a fast-growing field of research. Due to its capability, many useful applications can benefit from the deployment of a mobile robot, such as surveillance, emergency rescue, entertainment, library guides, medical care, industry collaboration and so on. Some of these applications involve human-robot interaction, in which a mobile robot must have abilities of perception, localization, navigation, locomotion and even cognition about people in its working environment. 
Person following \cite{islam2019person} is a capability that supports many useful applications of a mobile robot.

\begin{figure}[ht]
        \centering
        \subfigure[Height-based tracking with CCF \cite{koide2020monocular}]{
                \centering
                \label{fig:intro-baseline}
                \includegraphics[width=0.45\textwidth]{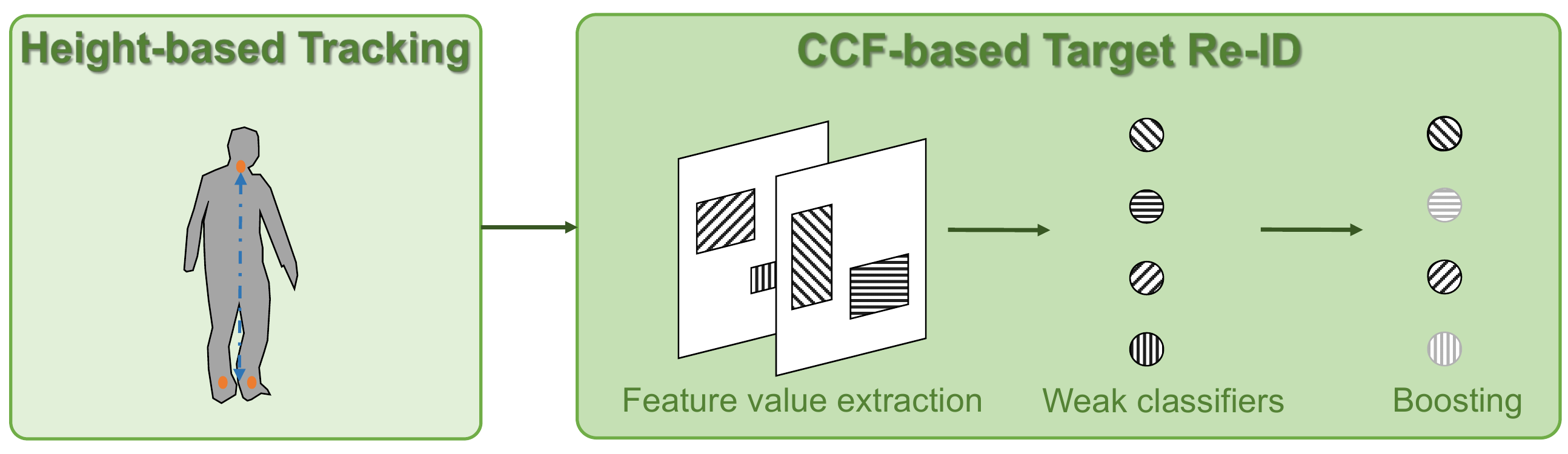}
        }
        \subfigure[Our width-based tracking with GRR_SLT]{
                \centering
                \label{fig:intro-ours}
                \includegraphics[width=0.45\textwidth]{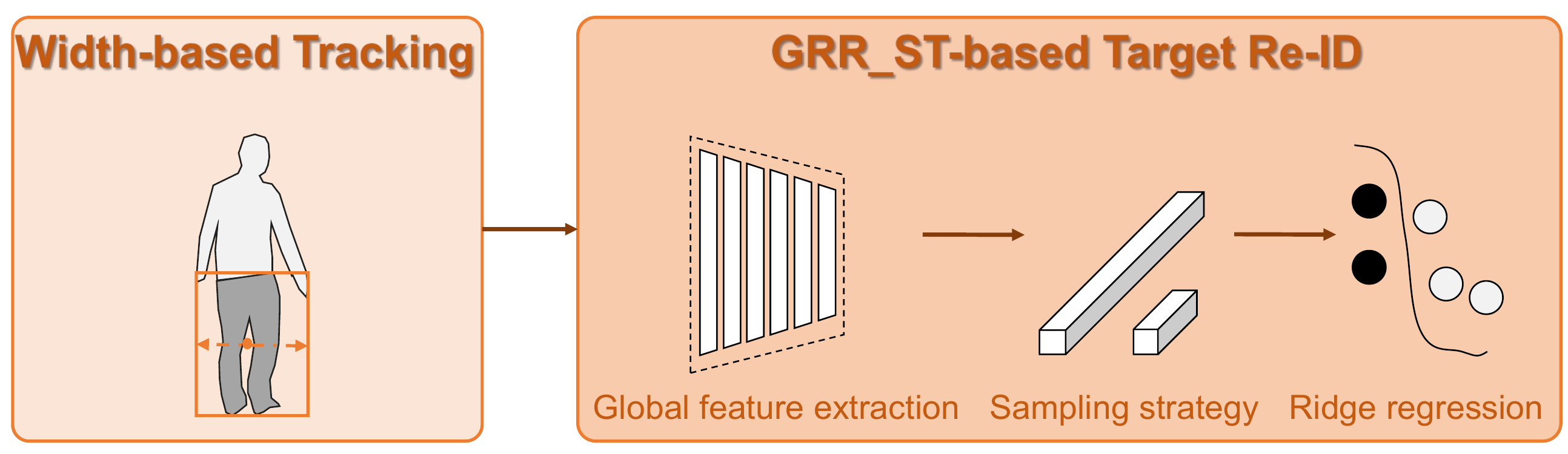}
        }
\caption{A MPF system consists of two key modules: tracking module and target Re-ID module. (a) is the SOTA method with a \textit{height-based} tracking module and a CCF (convolutional channel feature) based target Re-ID module. (b) is the proposed method with a \textit{width-based} tracking module and a GRR_SLT-based module (a global descriptor and a ridge regression combining with a short-long-term sample set)}.
\label{introduction}
\end{figure}

In order to perform person following, some proposed methods \cite{leigh2015person, sung2015hierarchical, yuan2018laser, wang2017real, chen2017integrating, koide2016identification, linder2016multi} track multiple people with the help of a distance measurement sensor such as LiDAR and RGBD camera. Once selecting one person in the field of view of the robot, it will follow the person based on the tracking position.
To deploy person following on low-cost mobile robots, \cite{zhang2019vision} uses a vision-based single object tracking (SOT) \cite{zheng2021improving, li2019siamrpn++, danelljan2019atom} to track the target, and relies on a visual servo to follow the target. 
\cite{koide2020monocular} proposes a monocular-vision person following (MPF) system to track and follow the target by an assumption of full-body observation.
However, these MPF systems still suffer from challenging situations involving close observation, target appearance change and highly similar appearance between the target and distracting people. 

To address the above problems, we propose a robust MPF system consisting mainly of a people tracking module and a target Re-ID module. 
Our people tracking module can obtain multiple people tracks even at a close distance to the target because our \textit{width-based} people detection and position estimation make use of the width information of people as a prior without requiring the full-body observation of people.
By using high-level global features of the target that are learned and adapted online, our target Re-ID module can re-identify the target even when the target is lost in difficult cases when it moves out of the view due to abrupt motion or distracting people of similar appearance appear in the scene. We rely on a sampling strategy to properly consider historic observations of the target used by the online classifier to mitigate an overfitting problem effectively.

\begin{figure*}[t]
        \centering
        \includegraphics[width=\textwidth]{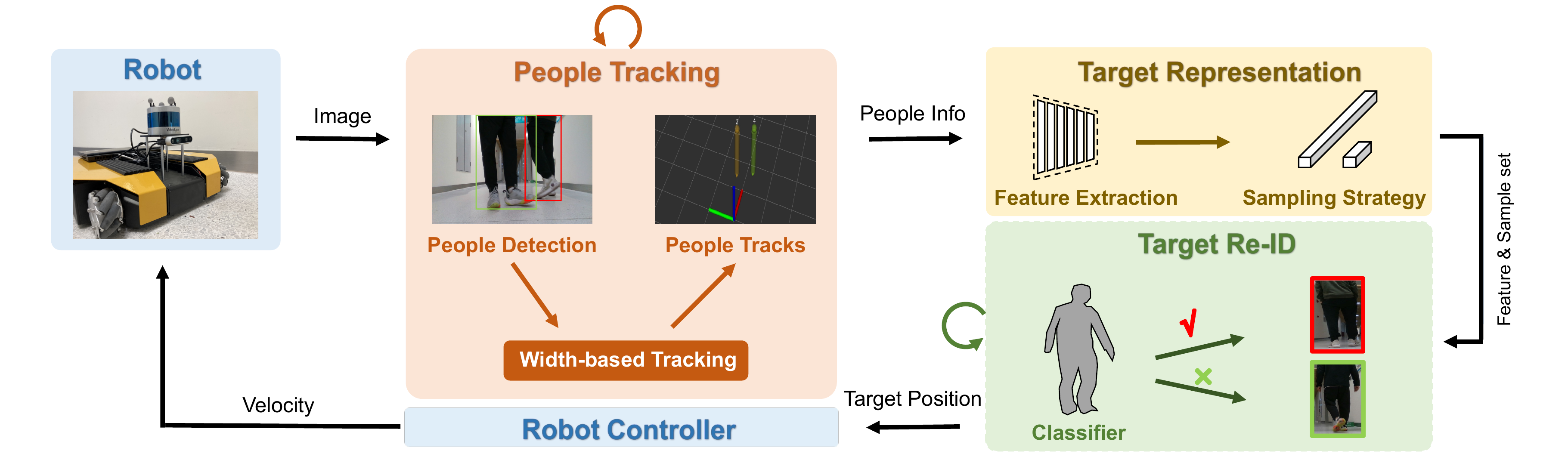}
        \caption{The framework of the MPF system, where the arrow with a circle line means previous information is used. Our main contributions are: 1) \textbf{\textit{width-based} tracking}, which can track people even at a close distance by utilizing the observale boxes; and 2) \textbf{target representation}, which is robust in challenging situations of distracting people of high similarity to the target by using high-level features and historic observations to construct a discriminative target appearance model.}
        \label{method-framework}
\end{figure*}


In summary, in this paper, we propose a robust MPF system  with the following two contributions:
\begin{itemize}
        \item{We design a \textit{width-based} people tracking module to endow robots with the ability to follow the target reliably even at close distance; and}
        \item {we propose a robust target Re-ID module considering high-level target features and historic observations in constructing a robust target appearance model.}
\end{itemize}

The rest of the paper is organized as follows. Section \ref{sec:related-works} presents related works and the motivation of our study. Section \ref{sec:methodology} introduces our MPF framework and details the proposed key modules involving people tracking module and target Re-ID module. Section \ref{sec:experiments-setup} is about experiments and implementation details. Section \ref{sec:results-discussion} shows the experimental results and discussion. Finally, section \ref{sec:conclusion} concludes this paper.

\section{RELATED WORKS} \label{sec:related-works}

\subsection{Monocular-Vision Person Following Robot}
Many existing works about the person following \cite{leigh2015person,sung2015hierarchical,yuan2018laser,wang2017real,chen2017integrating,koide2016identification,linder2016multi} use distance measurement sensors, which can be expensive and have difficulty in dealing with cluttered indoor environments. A few works are based on monocular vision. \cite{zhang2019vision} tracks people in the image space and realizes the person following by visual servo. For estimating a person's position in the robot space, inspired by person position estimation with the hypothesis of known height and ground plane position \cite{choi2010multiple, ardiyanto2014partial, hoiem2008putting}, \cite{koide2020monocular} proposes a \textit{height-based} MPF framework. However, it would fail at a close distance to the target because it requires that people's full bodies of people including their necks are observable. In real robots applications, the close-range following commonly occurs, e.g., in robot dog following and service robot collaboration. 
To solve this problem, we design a \textit{width-based} tracking method with a hypothesis of the known width of the target. 
Then we are able to track people at a close distance by using bounding boxes returned by a people detector that works reliably even at a close distance.

Target loss is a common occurrence in person following applications. To solve this problem, \cite{koide2020monocular} includes a target Re-ID module in its design as shown in Figure \ref{fig:intro-baseline}. The Re-ID module uses features that are computed from convolutional channel feature (CCF) map of a rectangle region. Then a number of weak Bayes classifiers are initialized by randomly choosing a CCF map and a rectangle region. Lastly, online boosting \cite{grabner2006line} is used to select weak classifiers to form a strong classifier. 
The main weakness of the proposed Re-ID module in [2] is that it cannot distinguish the target from distracting people of similar appearance due to the use of low-level features. To improve the Re-ID ability, we use a high-level global descriptor and a ridge regression model to learn a robust representation of the target online that is robust even in situations of high similarity of the target and distracting people.

\subsection{MOT, SOT and Sampling Strategy}

Much can be learned from the literature on multiple-object tracking (MOT) and single-object tracking (SOT) in order to design a learning strategy properly in our MPF problem.
MOT \cite{wojke2017simple, zhang2021fairmot, milan2016mot16} realizes multiple people tracking in the image space with boxes and pre-trained global descriptors. However, pre-trained descriptors are not adaptable, which would easily cause wrong Re-ID in situations of continuous appearance change of the target. Here, we combine the global descriptor with an online learning classifier to improve our target Re-ID ability by training with additional appearance samples generated through tracking.

SOT \cite{li2019siamrpn++, danelljan2019atom, mayer2021learning} tries to keep tracking a region of the target by searching for it in the neighborhood of the previously detected target region. It is a task including box regression and target classification. ATOM \cite{danelljan2019atom} first integrates online target classification learning into SOT to distinguish the target and its background. However, it cannot perform well when the target disappears and reappears. 
Here, we perform online target classification learning to distinguish the target and other distracting people instead of the target and the background.

In both ATOM \cite{danelljan2019atom} and exiting MPF works \cite{koide2020monocular}, the online classifier is trained by features computed from recent target observations. However, it will easily cause over-fitting, which would affect Re-ID performance in situations where appearance change frequently happens. The problem is also called \textit{catastrophic forgetting} \cite{parisi2019continual} in the deep learning literature. To alleviate it, \textit{experience replay} has been proposed by randomly adding historic samples to the recent sample set \cite{knowles2021toward, kuznietsov2021comoda, zhang2020online}. Inspired by this idea, we use a sampling strategy to construct a sample set consisting of the latest features and historic features of the target from past observations, to help build a training dataset that can effectively overcome the over-fitting problem.

\section{METHODOLOGY} \label{sec:methodology}

\subsection{Main Components of Our MPF}

The overall MPF framework is shown in Figure \ref{method-framework}. Its modules and their effects are as follows upon the selection of the target, a step whose solution is application-dependent:
\begin{enumerate}
\item \textbf{People Tracking:} Detect and track people based on the current measurements and the tracks of the last timestep.
\item \textbf{Target Representation:} Extract people's features using a feature extractor and construct a training sample set by a sampling strategy.
\item \textbf{Target Re-ID:} Classify and re-identify the target based on a target Re-ID logic; export the target position and train the classifier if the target is found.
\item \textbf{Robot Controller:} Control the mobile robot to follow the target based on the target position.
\end{enumerate}



\subsection{Width-based People Tracking} \label{sec:width-based-tracking}

In order to track people at a close distance, we design a \textit{width-based} tracking module. Such a module is superior to the \textit{height-based} people tracking because it can detect people and estimate their positions without requiring observation of the full body. Here, we use a Kalman filter to realize our \textit{width-based} tracking. Our method takes advantage of the assumption of a known body width of people, which can be easily satisfied in practice. 

Supposing extrinsic and intrinsic parameters are known as $\mathbf{R}_w^r,\mathbf{t}_w^r$ from world frame to robot frame, $\mathbf{R}_r^c,\mathbf{t}_r^c$ from robot frame to camera frame, and $f_x, c_x$ of camera intrinsic parameters. 
The raw measurement is defined as: $\mathbf{b}_k=[u_{tl}\ v_{tl}\ u_{br}\ v_{br}]^T$, where $tl$ and $br$ mean top-left and bottom-right point of the bounding box respectively.
Here, a person state in the world frame is defined as: $\mathbf{s}_k=[x_k\ y_k\ \dot{x_k}\ \dot{y_k}]^T$ consisting of position and velocity states. A constant velocity model is assumed to predict the state. 

For estimating the distance between the person and the camera, we make a hypothesis that the target person is a cylinder with $r$ radius in the direct front of the robot. We can then get the distance of a person $z^c$ in the camera frame (detailed derivation and discussion are provided in Appendix \ref{sec:distance-estimation}):
\begin{equation}
z^c=f_x\cdot{\frac{r}{u_{br}-u_{tl}}}
\label{distance-estimation}
\end{equation}

Then supposing $\mathbf{\bar{s}}=[x\ y\ z]^T$, combining with Equation \ref{distance-estimation}, and according to ridgy body transformation and projective transformation, we can get the observation equations that relate the box bounding variables of a tracked person and the person's position as follows.:
\begin{subequations}
\begin{align*}
        \label{equ:observation-euqation-a}
        f_x\cdot{\frac{(\mathbf{R}_r^c(\mathbf{R}_w^r \mathbf{\bar{s}} + \mathbf{t}_w^r) + \mathbf{t}_r^c)|_x}{(\mathbf{R}_r^c(\mathbf{R}_w^r \mathbf{\bar{s}} + \mathbf{t}_w^r) + \mathbf{t}_r^c)|_z}}+c_x =\frac{u_{tl}+u_{br}}{2} \tag{2a}  \\
        \label{equ:observation-euqation-b}
        (\mathbf{R}_r^c(\mathbf{R}_w^r \mathbf{\bar{s}} + \mathbf{t}_w^r) + \mathbf{t}_r^c)|_z = f_x\cdot{\frac{r}{u_{br}-u_{tl}}}, \tag{2b}
\end{align*}
\end{subequations}
where $|_x$ means the $x$ value of the point. $|_z$ is similar.

This derivation results in a linear observation model, whose details are shown in Appendix \ref{sec:observation-model}. 
Supposing the expected obervation is $\mathbf{o}_k$.
Through the linear observation model, we can get $\mathbf{o}_k$. 
To establish the data association between $\mathbf{o}_k$ and a raw measurement $\mathbf{b}_k$, we need to change $\mathbf{b}_k$ to the form as $\mathbf{o}_k$. Therefore, our \textit{processed measurement} $\mathbf{y}_k$ is as follows:
\begin{equation}
\mathbf{y}_k =
        \begin{bmatrix}
                \frac{r\cdot(u_{tl,k}+u_{br,k}-2c_x)}{2(u_{br,k}-u_{tl,k})}-(\mathbf{t}_r^c|_x)^2-[1\ 0\ 0]\mathbf{R}_r^c \mathbf{t}_w^r\\
                {\frac{f_{x}\cdot{r}}{u_{br,k}-u_{tl,k}}}-(\mathbf{t}_r^c|_z)^2-[0\ 0\ 1]\mathbf{R}_r^c \mathbf{t}_w^r
        \end{bmatrix},
\label{equ:pro-measurement}
\end{equation}
where only bounding box information is required to be measured. Thus, we can obtain the measurements at a close distance without requiring full-body observation.

Due to inaccuracy of the detected bounding boxes when two people overlap, we keep bounding box of a person only if its largest IoU with other boxes is smaller than a threshold ${\delta}_{iou}$.
\begin{subequations}
\begin{align*}
\mathbf{B}_k = \{\mathbf{x}_i | f(\mathbf{x}_i, \mathbf{\bar{B}}_k)<{\delta}_{iou},\mathbf{x_i} \in \mathbf{B}_k, \mathbf{\bar{B}_k}=\mathbf{B}_k \setminus{\{\mathbf{x}_i\}}\} \tag{4a} \\
f(\mathbf{c}, \mathbf{Q}) = \mathop{\max}\limits_{\mathbf{q}_i \in \mathbf{Q}}IoU(\mathbf{c}, \mathbf{q}_i), \tag{4b}
\label{equ:filter-measurements}
\end{align*}
\end{subequations}
where $\mathbf{B}_k$ is the set of raw measurements at $k$ timestep.

After that, we can use Equation \ref{equ:pro-measurement} to get the \textit{processed measurement}. Here, we calculate the Euclidean distance between the \textit{processed measurement} and the expected observation, so our distance metric is:
\begin{equation}
        d(i, j) = ||\mathbf{o}^i - \mathbf{y}^j||_2^2
\label{equ:distance-metric}
\end{equation}

Then, a GNN (global nearest neighbor) is used to match the processed measurements to the predicted Kalman states. After Kalman updates, we can get updated people's states.

The tracks information is then added to the people information for target Re-ID. Besides, the people information also contains corresponding image patches and boxes information.

\subsection{Target Representation and Classifier} \label{sampling-strategy}

With the bounding boxes of detected people in the current view of the robot camera, we first extract their features by a pre-trained CNN. Here, we choose to use a global descriptor as in DeepSORT \cite{wojke2017simple} in order to overcome the weakness of a local descriptor such as \cite{koide2020monocular} so that our target Re-ID module can handle distracting people of similar appearance to our target.

In addition, we adopt the online learning module in \cite{koide2020monocular} in order to handle the continuous changes of the target appearance with respect to viewpoint and lighting conditions as well as to take advantage of the additional appearance samples generated through tracking. Instead of the Bayes classifier used in \cite{koide2020monocular}, we use a ridge regression model with L2 regularization as our online learning classifier. Such a regularization-based classifier is able to alleviate the over-fitting problem caused by the limited numbers of the training set. 

In the meantime, inspired by \textit{experience replay} proposed in \cite{knowles2021toward, kuznietsov2021comoda, zhang2020online}, we construct a training sample set containing the latest features and historic features to mitigate the over-fitting problem caused by the lack of diversity in the latest samples. Historic features are selected based on the people information from the tracking module.

\subsection{Target Re-ID and Robot Controller}

For a complete MPF system, we also need to provide the Re-ID logic and the robot controller here. Our Re-ID logic is mainly based on \cite{koide2020monocular}. In every frame, the classifier would predict the score of the target. If the score is lower than a threshold $\delta_{switch}$, then the system will turn to \textit{Re-ID} state for judging an id-switch is happening. If the target id is lost, it will also lead to \textit{Re-ID} state. 
In \textit{Re-ID} state, all candidates will be predicted by the classifier. The candidate will be judged as the target if its predicted score is larger than a threshold $\delta_{id}$ in $N_{id}$ consecutive frames.

Subsequently, a proportional-integral-derivative controller is used for the robot control. Specifically, in the robot frame, we control the robot by maintaining a given $x$ value and reducing $y$ value to be zero for stable distance estimation by Equation \ref{distance-estimation}.




\section{EXPERIMENTS SETUP} \label{sec:experiments-setup}

\subsection{Datasets and Evaluation Metrics}

Here, we use three datasets in the experiments. One is for the evaluation of our \textit{width-based} tracking module, and the others two are used for the target Re-ID evaluation of the whole MPF system. The first dataset consists of sequential frames and the poses of the target person and the following robot in every frame whose poses are collected by a motion capture (MC) system. Five sequences are collected by walking in the front of the robot within 0.5 m - 7.0 m to allow us to improve the repeatability of the experimental results.

Two other datasets consist of only image sequences. One is a public person following dataset \cite{chen2017integrating}. It contains 11 sequences that are captured by a stereo camera with challenging target Re-ID situations involving illumination change and clothes change.
Another dataset is a custom-built dataset, which is designed to fill the gap of the public dataset, for its lack of challenging situations including long-term people occlusion, frequent distance change and similar clothes. It contains four sequences named as \textit{corridor1, corridor2, lab_corridor} and \textit{room}. 

\begin{figure}[t]
\centering
\subfigure[corridor1]{
        \centering
        \includegraphics[width=0.45\textwidth]{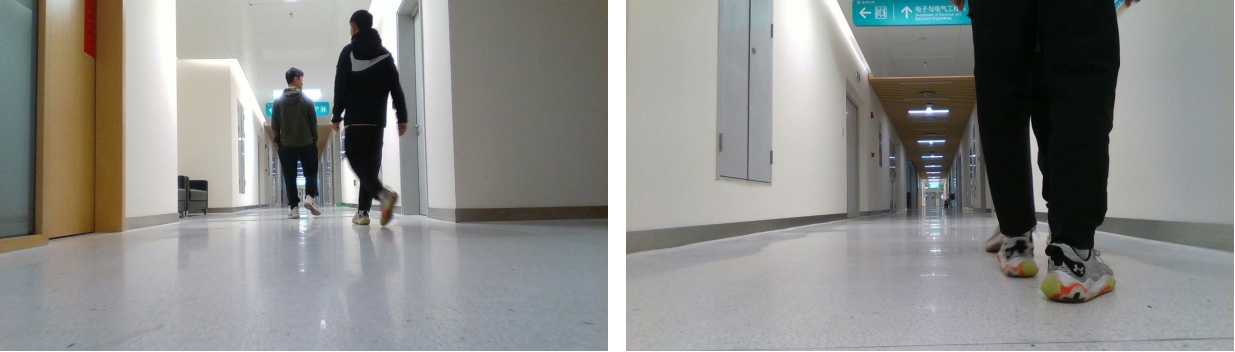}
}
\subfigure[room]{
        \centering
        \includegraphics[width=0.45\textwidth]{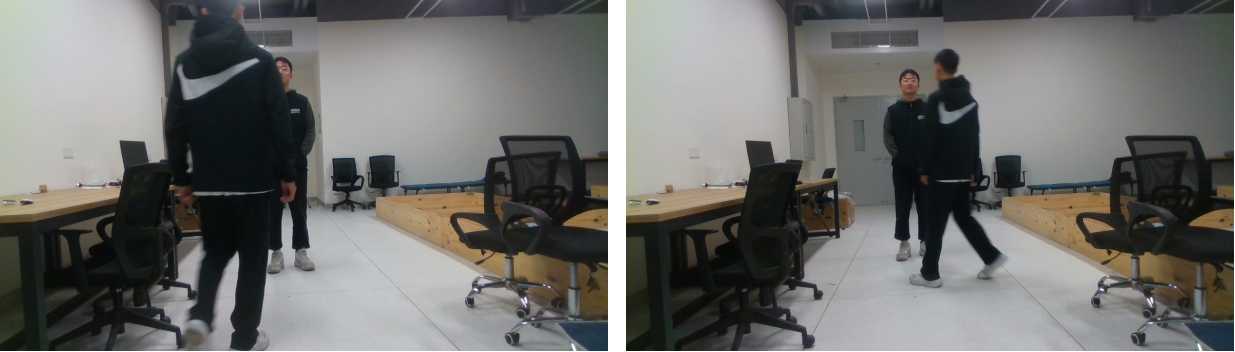}
}
\caption{Examples of a custom-built dataset with challenging situations including long-term people occlusion, frequent distance change and similar clothes.}
\label{datasets}
\end{figure}

Their attributes and corresponding degrees are listed in Table \ref{datasets-attributes} and some examples of these sequences are shown in Figure \ref{datasets}.  
\textit{corridor1}, \textit{corridor2} and \textit{lab_corridor} are with dissimilar appearance of upper bodies and similar lower bodies, while \textit{room} is captured with totally similar appearance. 
\textit{corridor2} and \textit{lab_corridor} have only one long-term occlusion, and \textit{corridor1} and \textit{room} have two times, but the occlussion of \textit{corridor1} is more serious. For short-term occlusion, mutual crossing exists in \textit{corridor2} and \textit{lab_corridor}, a situation that does not occur in \textit{corridor1} and \textit{room}. In addition, distance change occurs in all sequences.

In the last two datasets, we evaluate the Re-ID capability in terms of accuracy of target person localiztion in the image space. In each frame, if the distance between the center of the ground truth box and the center of the estimated target person region is smaller than a threshold, we regard Re-ID as being successful.

\begin{table}[t]
        \caption{Attributes of the custom-built dataset. More $+$ means greater degree, and $-$ means that this attribute is not involved in the dataset.}
        \vspace{5pt}
        \centering
        \scalebox{0.8}{
                \begin{tabular}{ccccc}
                        \toprule
                                &\textbf{\textit{corridor1}} &\textbf{\textit{corridor2}} &\textbf{\textit{lab_corridor}} &\textbf{\textit{room}} \\\midrule
                        Similarity &+ &+ &+ &++ \\ 
                        Long-term occlusion &+++ &+ &+ &++ \\ 
                        Short-term occlusion &- &++ &++ &- \\
                        Distance change&+ &++ &+ &+ \\
                        \bottomrule
        \end{tabular}}
        \label{datasets-attributes}
\end{table}

\subsection{Baselines and Our Method}
In the public dataset, we compare the proposed method with the SOTA method \cite{koide2020monocular} named as \textit{HEIGHT_CCF} consisting of a \textit{height-based} tracking module and a \textit{CCF} Re-ID module, and other methods reported in \cite{chen2017integrating} including \textit{OAB} \cite{grabner2006real}, \textit{ASE} \cite{danelljan2014accurate}, \textit{SOAB} \cite{chen2017person}, \textit{DS-KCF} \cite{camplani2015real}, \textit{CNN_v1}, \textit{CNN_v2} and \textit{CNN_v3} \cite{chen2017integrating}. These reported methods are SOT-based methods (\textit{OAB} and \textit{ASE}) and SOT-based methods combining with stereo camera (\textit{SOAB} and \textit{DS-KCF}) or RGBD camera (\textit{CNN_v1}, \textit{CNN_v2} and \textit{CNN_v3}). 
To compare \textit{height-based} and \textit{width-based} tracking method in a fair way, we evaluate a method called \textit{WIDTH_CCF}, which consists of a \textit{width-based} tracking module and a CCF Re-ID module. While the proposed \textit{GRR} is involved in \textit{WIDTH_GRR}.

The custom-built dataset is used to compare the effectiveness of the Re-ID modules including \textit{CCF} (SOTA) and ours. The proposed Re-ID method and the \textit{CCF} are evaluated combined with a \textit{width-based} tracking module for a fair comparison. Furthermore, for revealing the influence of the size of the sample set on the effectiveness of the online learning Re-ID module, we make a study in terms of different sizes of the sample set. So \textit{CCF} modules with 16, 32, 64 and 128 sizes of the sample set are named as \textit{CCF_16}, \textit{CCF_32}, \textit{CCF_64} and \textit{CCF_128} respectively. Similarly, \textit{GRR_ST_16}, \textit{GRR_ST_32}, \textit{GRR_ST_64} and \textit{GRR_ST_128} are corresponding to the different sizes setting of \textit{GRR} with short-term sample set. The proposed method with both the long-term and short-term samples is named as \textit{GRR_SLT_64}, which consists of \textit{GRR} and the sampling strategy.

\subsection{Implementation Details}
Our people detection model is YOLOX \cite{ge2021yolox} and feature extraction model$\footnote{https://github.com/pmj110119/YOLOX_deepsort_tracker}$ is similar to DeepSORT \cite{wojke2017simple} whose global descriptor has a dimension of 512. In people tracking, ${\delta}_{iou}=0.5$. In target identification, $N_{id}=5$, ${\delta}_{switch}=0.35$ and ${\delta}_{id}=0.60$.

A Clearpath Dingo-O, a Realsense D435i with $1280\times720$ and 30Hz, and a laptop with Intel(R) Core(TM) i5-10200H CPU @ 2.40GHz and NVIDIA GeForce RTX 1650 are used in the person following procedure. 
All the datasets stored in rosbag format on a computer with Intel® Core™ i7-10700F CPU @ 2.90GHz and NVIDIA GeForce RTX 2060.

\section{RESULTS \& DISCUSSION} \label{sec:results-discussion}
\subsection{Effectiveness of Width-based Person Following}
\begin{figure}[t]
        \centering
        \includegraphics[width=0.47\textwidth]{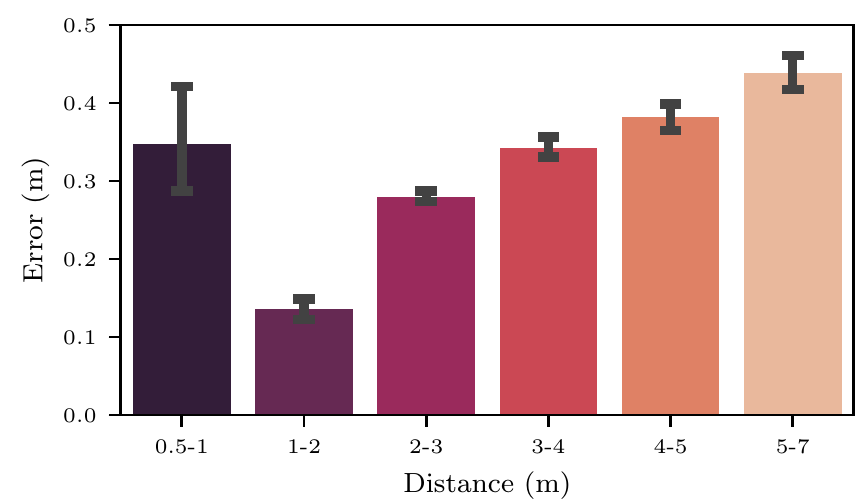}
        \caption{Box plot of error respect to distance, where distance is the ground truth distance from the target to the robot captured by our MC system, and the error come from the ground truth distance and the estimated distance. In 0.5-1 m, tracking mean error is almost 0.35 m. From 1 m to 7 m, the error is getting larger.}
        \label{tracking-accuracy-distribution}
\end{figure}

\begin{figure}[t]
        \centering
        \includegraphics[width=0.47\textwidth]{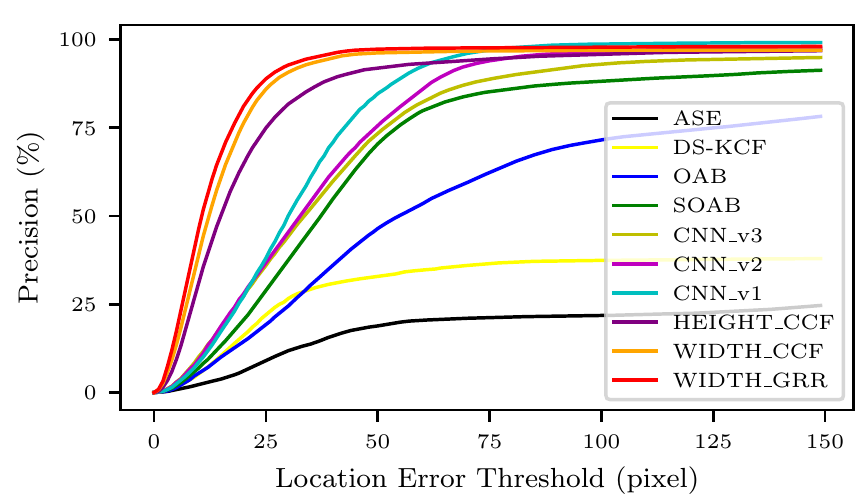}
        \caption{Plot of precision respect to different location error thresholds. \textit{WIDTH_CCF} performs better than \textit{HEIGHT_CCF}, which indicates \textit{width-based} tracking could achieve better performance.}
        \label{trackingPlot}
\end{figure}

\begin{figure}[t]
        \centering
        \includegraphics[width=0.47\textwidth]{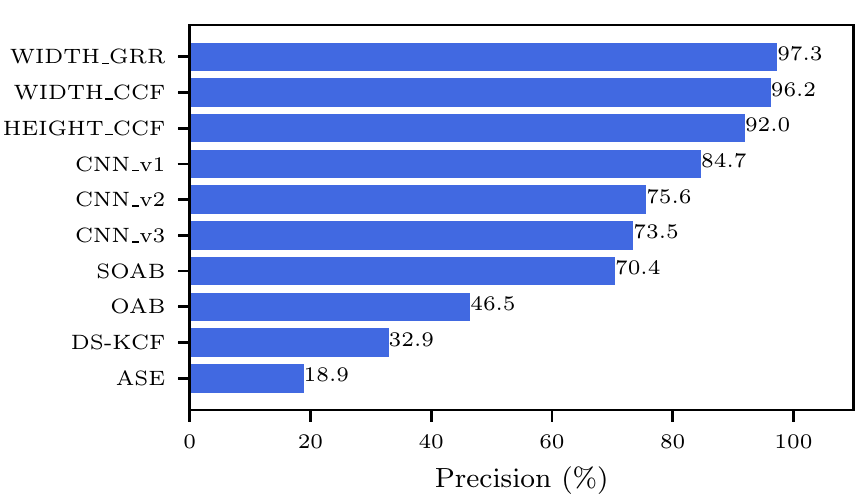}
        \caption{Precision in location error threshold of 50 pixels. \textit{WIDTH_CCF} is better than \textit{HEIGHT_CCF} with 96.2\% vs. 92.0\%, and our proposed method \textit{WIDTH_GRR} achieves the best result of 97.3\% precision.}
        \label{trackingBar}
\end{figure}

\begin{figure}[t]
        \subfigure[]{
                \centering
                \includegraphics[width=0.14\textwidth]{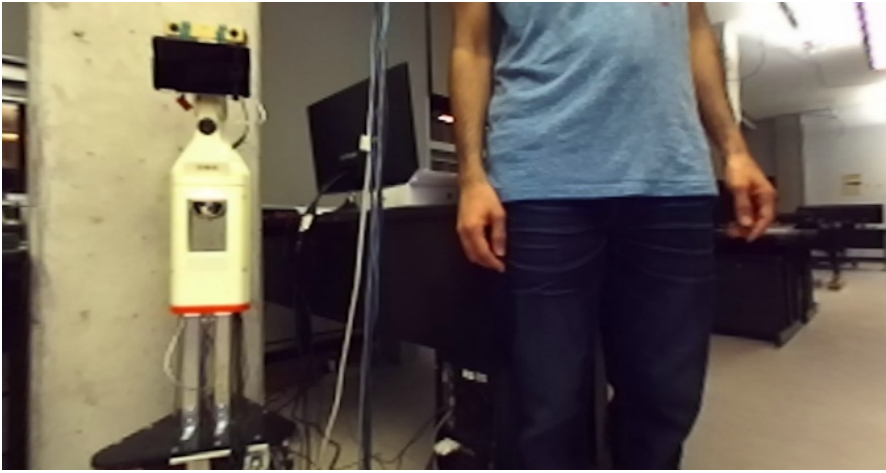}
        }
        \subfigure[]{
                \centering
                \includegraphics[width=0.14\textwidth]{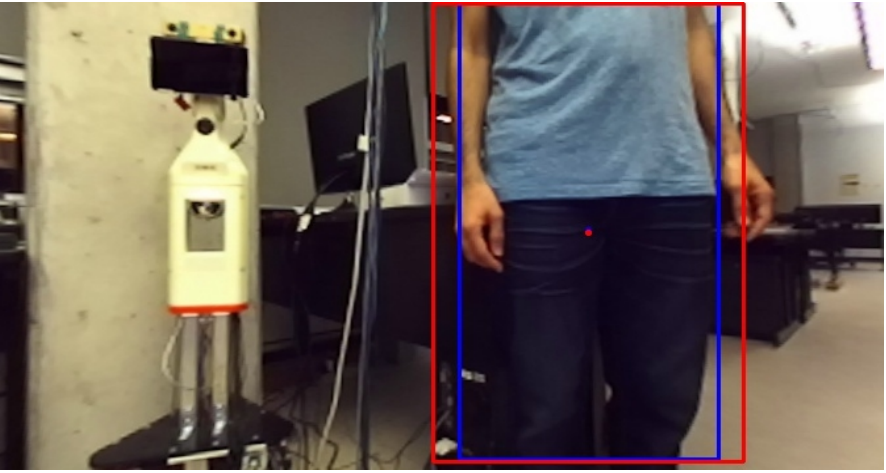}
        }
        \subfigure[]{
                \centering
                \includegraphics[width=0.14\textwidth]{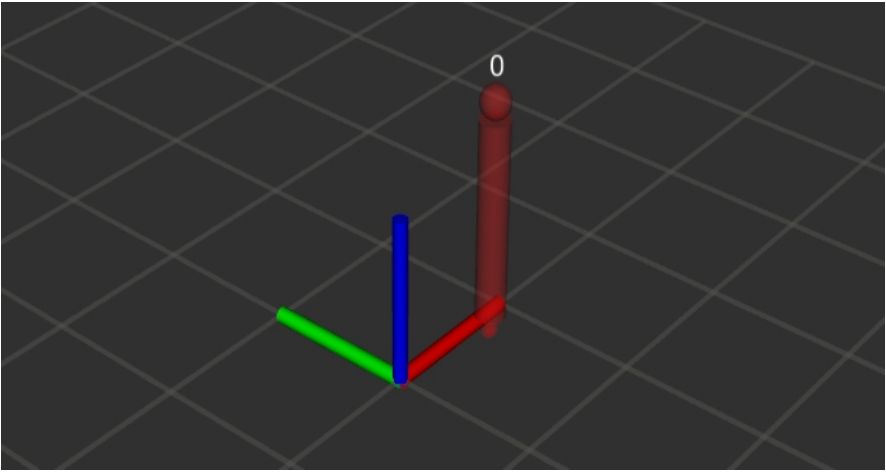}
        }
        \caption{(a) is the failed case of \textit{HEIGHT_CCF} for its failed observation of full body at close distance. (b) is our result with a red box and the ground truth with a blue box. (c) is the result of our method in the robot frame.}
        \label{trackingPic}
\end{figure}

\subsubsection{Accuracy of our width-based tracking module}
As shown in Figure \ref{tracking-accuracy-distribution}, the total estimation error of our tracking module is smaller than 0.5 m, which is appropriate for the MPF.
The mean error when the following distance is between 0.5 and 1 meter is larger than  that when the following distance is between 1 and 2 meters, 2 and 3 meters, and 3 and 4 meters, and its variance is larger than all other situations. 
This phenomenon could be attributed to the difference between the real person and our hypothetical model of a person. 
We assume that the person is a cylinder with radius $r$, where $r$ is measured by the upper body; however, the radius of the upper body is different from the lower body. 
Thus, our distance estimation method could be further improved by a more precise person model, but the error below 0.5 m is enough for a simple application.
When the following distance between 1 and 7 meters, the error gets larger as the distance gets larger. 
This could be attributed to the bias of our distance estimation, where the target is assumed to be directly in front of the robot, but our walking direction is random in the experiment. 
So the bias would be larger as the distance gets larger based on the projection theory.
But this bias can be mitigated by our controller.

\subsubsection{Effectiveness of our width-based tracking module}
From Figure \ref{trackingPlot}, we can observe that \textit{WIDTH_CCF} surpasses \textit{HEIGHT_CCF} and other reported methods with a larger precision.
As shown in Figure \ref{trackingBar}, the precision of \textit{WIDTH_CCF} is 96.2\% versus 92.0\% of \textit{HEIGHT_CCF} at 50 pixel threshold. 
This result is mainly attributed to the effect of the \textit{width-based} tracking method. An example is shown in Figure \ref{trackingPic}.
The left image is the image which is failed to be detected and tracked by \textit{HEIGHT_CCF} without observation of the person's full body or neck.
The middle image is our detection result with a red box and the ground truth with a blue box.
The right image is our tracking result of the target person in the robot frame. 
Thus, such a good performance in the public dataset depends to a large extent on the \textit{width-based} tracking, which can utilize the boxes of the people to realize the people tracking even at a close distance.
From the above observation and analysis, we can conclude that our \textit{width-based} tracking module is beneficial for the MPF in the situations of distance change for its successful tracking even at a close distance.

And our \textit{WIDTH_GRR} is a little better than \textit{WIDTH_CCF} in terms of 97.3\% versus 96.2\% for its better Re-ID capability in situations of distance change between the people and the robot, which would be further discussed in the following.

\subsection{Performance of our Re-ID Module}
\subsubsection{CCF vs. GRR_ST}
Here, we compare the Re-ID ability of \textit{CCF} and \textit{GRR_ST} on our dataset and the best results of them are selected to be compared. From Table \ref{Re-ID-precission}, we can see that \textit{GRR_ST} outperform \textit{CCF} in all sequences, by 6.4\% in \textit{corridor1}, 9.4\% in \textit{corridor2}, 2.2\% in \textit{lab_corridor} and 62.7\% in \textit{room}. 
We can observe that high similarity is fatal to \textit{CCF} for it only gets 34.5\% mAP in \textit{room}, where the target is wrongly re-identified when the first occlusion happened. 
Oppositely, \textit{GRR_ST} acts well after two long-term occlusions. 
\textit{CCF} performs badly because it only uses a sum value from a region of a low-level feature map as its feature value, which would cause confusion for the Naive Bayes classifier in the situations of high similarity. 
While \textit{GRR_ST} uses a global descriptor as its feature vector, which not only integrates appearance information but also contains the spatial relation information of a person's parts, leading to a more discriminative representation of the target.
So the classifier is able to distinguish the target from distracting people in high similarity cases.

In \textit{corridor2} and \textit{lab_corridor}, both \textit{CCF} and \textit{GRR_ST} perform well, where \textit{GRR_ST} is a little better with 98.4\% vs. 89.0\% of \textit{CCF} in \textit{corridor2} and 92.7\% vs. 84.4\% in \textit{lab_corridor}.
\textit{GRR_ST} can find the target once the target occurs without any hesitation, while \textit{CCF} can find the target again after occlusion until the discriminative upper body appeared in the image. 

In \textit{corridor1} in which long-term occlusion is the most severe, both methods act well after the first occlusion for the reason that the discriminative part of the body could be observed. 
But both of them are failed to re-identify the target in the second occlusion. Before the occlusion happened, only lower body parts that are not differentiated could be observed and this process lasts for a long time. So the sample set is full of these confusing samples, which leads to an over-fitting problem.

In conclusion, compared to \textit{CCF}, \textit{GRR_ST} is more discriminative for its better Re-ID ability in \textit{room} of high similarity, and faster Re-ID speed in situations of distance change (\textit{corridor2} and \textit{lab_corridor}). From the above analysis, we can conclude that combining with a high-level global descriptor can help to improve the robustness of target Re-ID for its superior feature representation ability.

\begin{table}[t]
	\caption{Average precision (\%) of the baseline and our method in the custom-built dataset. }
	\vspace{5pt}
	\centering
	\scalebox{0.9}{
		\begin{tabular}{ccccc}
			\toprule
			 &\textbf{\textit{corridor1}} &\textbf{\textit{corridor2}} &\textbf{\textit{lab_corridor}} &\textbf{\textit{room}} \\ \midrule
			CCF_16 &55.0 &89.0 &32.6 &32.7 \\
			CCF_32 &53.4 &80.6 &90.5 &32.6 \\ 
			CCF_64 &41.2 &39.7 &83.0 &33.0 \\
			CCF_128&53.9 &49.9 &84.4 &34.5 \\ \midrule
			GRR_ST_16&61.4 &32.3 &89.7 &97.0 \\
			GRR_ST_32&61.4 &37.7 &90.6 &97.2 \\ 
			GRR_ST_64&61.4 &98.3 &90.6 &97.2 \\ 
			GRR_ST_128&61.0 &\textbf{98.4} &92.7 &35.4 \\ \midrule
			GRR_SLT_64&\textbf{99.3} &95.6 &\textbf{92.8} &\textbf{97.2} \\
			\bottomrule
	\end{tabular}}
	\label{Re-ID-precission}
\end{table}

\subsubsection{GRR_ST vs. GRR_SLT}
From the sample set size study of \textit{GRR_ST} in Table \ref{Re-ID-precission}, we can observe that the online training of a classifier is sensitive to the sample set. 
In \textit{corridor2} of severe distance change, \textit{GRR_ST} with only the latest samples (16 and 32 samples) perform badly, which could be attributed to the over-fitting problem.
In \textit{room}, the classifier performs poorly when too many old samples (128 samples) are added, while it performs well with the latest samples (16, 32 and 64 samples).
These phenomenons indicate that historic observations could be contributed to the classifier.
But it's important to answer \textit{how historic} is the long-term samples and \textit{how to select} them.
\textit{Experience replay} \cite{knowles2021toward, kuznietsov2021comoda, zhang2020online} randomly select historic samples from previously-seen examples from a ``large dataset" that is different from the current samples.
Our sampling strategy, as is mentioned in Section \ref{sampling-strategy}, is similar to them. Our ``large dataset" is built by the selection of historic observations instead of a large sample set that contains the latest observations.

With the proposed sampling strategy, we can attain 99.3\% AP in \textit{corridor1}, 92.8\% AP in \textit{lab_corridor} and 97.2\% AP in \textit{room} with a size of 64. 
The outstanding performance in \textit{corridor_1} could be attributed to its historical memory, which can help to get rid of the over-fitting problem caused by the latest observations of the indiscernibility of the lower bodies. But it achieves 95.6\% AP in \textit{corridor2} vs. 98.3\% AP of \textit{GRR_ST_64}, where the Re-ID speed of \textit{GRR_SLT_64} is slower. 
This means that our sampling strategy can be further improved.

Overall, our sampling strategy can help the classifier alleviate the over-fitting effects by adding historic observations. Similar to the effect of a global descriptor, samples with diversity can also help to construct a high-level representation of the target, which is useful for target Re-ID.

\section{CONCLUSION} \label{sec:conclusion}

In this paper, we propose a MPF system with a \textit{width-based} tracking module and a robust target Re-ID module.
The results of experiments about the \textit{width-based} tracking module indicate that this module is accurate for the person following with an overall error lower than 0.5m.
And most importantly, it can track the target even at a close distance because our \textit{width-based} tracking module can track people without requiring full-body observation.

Also, our method achieves the best results in a custom-built dataset, which is beneficial from the discriminative ability of a high-level global descriptor.
Besides, the historic samples selected by the sampling strategy also help to describe the target at a high-level representation to alleviate the over-fitting problem.

In the future, we will further improve the target Re-ID ability of our system by integrating the graph information of the target and the people or body parts of the target.



\section*{APPENDIX}

\subsection{Distance Estimation} \label{sec:distance-estimation}
Human boxes coordinates are defined by two endpoints: $(u_{tl},v_{tl}),(u_{br},v_{br})$, and the corresponding points in the camera frame are $(x^c_{tl},y^c_{tl},z^c_{tl})$ and $(x^c_{br},y^c_{br},z^c_{br})$ respectively. And we suppose the target is in the directly front of the camera (this can be realized by the robot controller module), so we have $|x^c_{tl}-x^c_{br}|=r$ and $z^c_{tl}=z^c_{br}=z^c$. With camera projection equations:
\begin{subequations}
\begin{align*}
        \label{equ:6a}
        u_{tl} = f_x\cdot{\frac{x^c_{tl}}{z^c_{tl}}}+c_x \tag{6a}\\
        \label{equ:6b}
        u_{br} = f_x\cdot{\frac{x^c_{br}}{z^c_{br}}}+c_x \tag{6b},
\end{align*}
\end{subequations}
then subtracting Equation \ref{equ:6a} to Equation \ref{equ:6b}, and substituting above conditions, finally we can get Equation \ref{distance-estimation}.

\subsection{Linear Observation Model} \label{sec:observation-model}
Assembling Equation \ref{equ:observation-euqation-a} to Equation \ref{equ:observation-euqation-b}, and multiplying out, we can get:
\begin{subequations}
\begin{align*}
        (\mathbf{R}_r^c \mathbf{R}_w^r \mathbf{\bar{s}} + \mathbf{R}_r^c \mathbf{t}_w^r)|_x + (\mathbf{t}_r^c|_x)^2 =\frac{|u_{tl}+u_{br}|/2-c_x}{f_x} \tag{7a}  \\
        (\mathbf{R}_r^c \mathbf{R}_w^r \mathbf{\bar{s}} + \mathbf{R}_r^c \mathbf{t}_w^r)|_z + (\mathbf{t}_r^c|_z)^2 = f_x\cdot{\frac{r}{|u_{tl}-u_{br}|}} \tag{7b},
\end{align*}
\end{subequations}
according to the hypothesis and definitions in Section \ref{sec:width-based-tracking}, setting $z=0$ (supposing the person is a point on the x-y plane), we have:
\begin{equation}
\begin{aligned}
        &\begin{bmatrix}
                r_{00}& r_{01}\\ 
                r_{20}& r_{21}
        \end{bmatrix}\begin{bmatrix}
                x\\
                y
        \end{bmatrix}\\
        = &\begin{bmatrix}
                \frac{r\cdot(u_{tl}+u_{br}-2c_x)}{2(u_{br}-u_{tl})}-(\mathbf{t}_r^c|_x)^2-[1\ 0\ 0]\mathbf{R}_r^c \mathbf{t}_w^r\\
                {\frac{f_{x}\cdot{r}}{u_{br}-u_{tl}}}-(\mathbf{t}_r^c|_z)^2-[0\ 0\ 1]\mathbf{R}_r^c \mathbf{t}_w^r,
        \end{bmatrix},
\end{aligned}
\end{equation}
change $[x\ y]^T$ to $\mathbf{s} = [x\ y\ \dot{x}\ \dot{y}]^T$, fill $0$ to the left matrix, and define $\mathbf{R}_w^r \mathbf{R}_r^c=[r_{00}\ r_{01}\ r_{02};\ r_{10}\ r_{11}\ r_{12};\ r_{20}\ r_{21}\ r_{22}]$, then we can get our linear observation model:
\begin{equation}
\begin{aligned}
        &\begin{bmatrix}
                r_{00}& r_{01}& 0& 0\\ 
                r_{20}& r_{21}& 0& 0
        \end{bmatrix}\mathbf{s}\\
        = &\begin{bmatrix}
                \frac{r\cdot(u_{tl}+u_{br}-2c_x)}{2(u_{br}-u_{tl})}-(\mathbf{t}_r^c|_x)^2-[1\ 0\ 0]\mathbf{R}_r^c \mathbf{t}_w^r\\
                {\frac{f_{x}\cdot{r}}{u_{br}-u_{tl}}}-(\mathbf{t}_r^c|_z)^2-[0\ 0\ 1]\mathbf{R}_r^c \mathbf{t}_w^r,
        \end{bmatrix}
\end{aligned}
\end{equation}

\bibliographystyle{IEEEtran}
\bibliography{ref}
\end{document}